%% file: results.tex
\section{Performance Evaluation}
\label{section:performance}

\subsection{Methodology}
\label{section:methodology}

\begin{figure}[t]
      \centering
      \includegraphics[width=0.90\columnwidth]{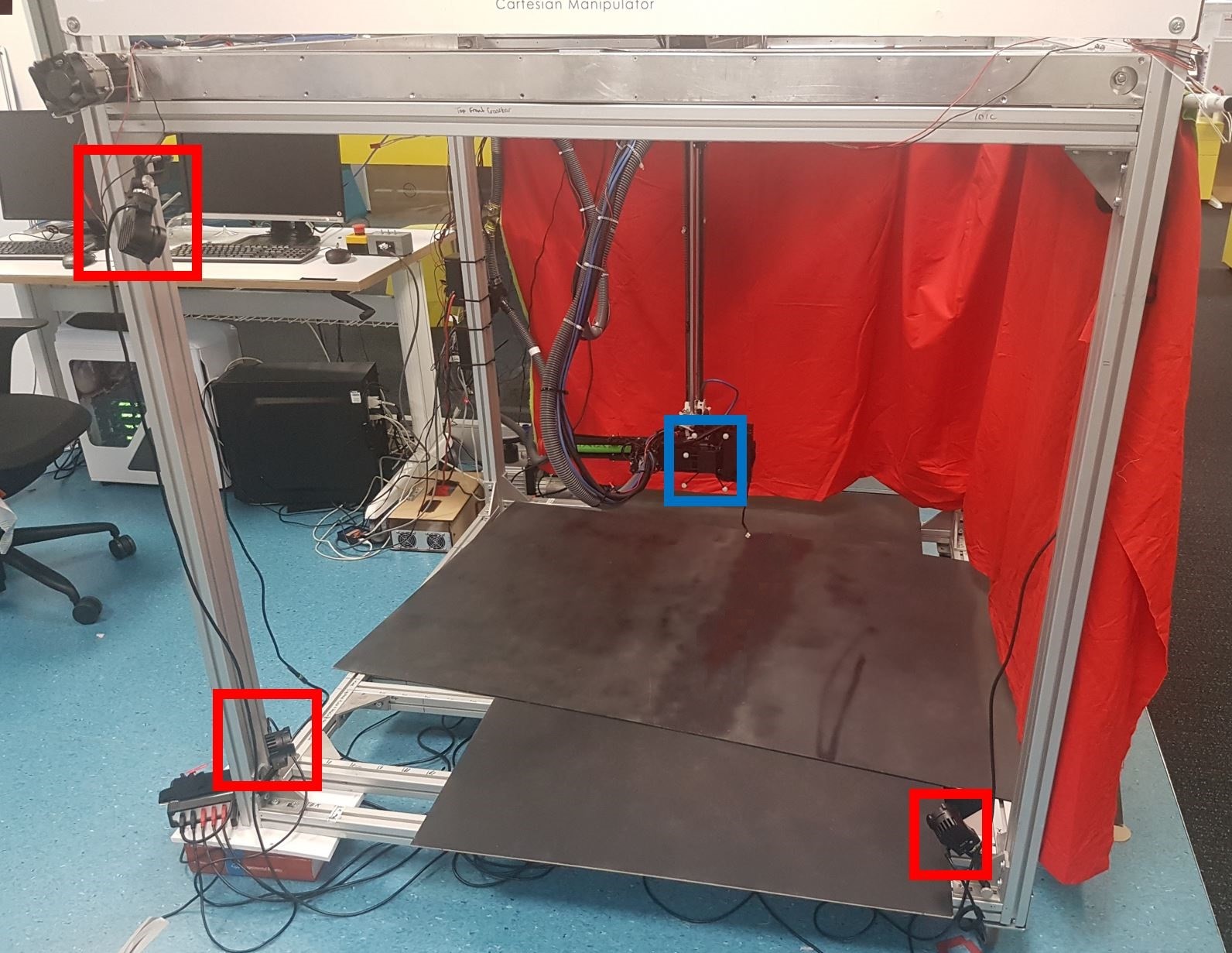}
      \caption{Motion capture setup. Five motion capture cameras (red boxes) were used to determine the pose of the multi-reflective body (blue box) }
      \label{fig:motive_setup}
\end{figure}

In order to test whether the design specs have been met, the motion of the system was tested. OptiTrack Motive, a motion capture system was used to track trajectories during different path executions. The setup can be seen in Fig.~\ref{fig:motive_setup}. Each axis was tested individually at various heights and speeds to test the end point accuracy during motion. Each axis was commanded at a fast and slow speed moving to and from the extremes of the tracking system. The pose of the multi-reflective body seen in Fig.~\ref{fig:motive_setup} is tracked and interpolated to compare the desired joint position with recorded position whilst in motion and for steady state error.\\

\begin{figure}[]
      \centering
      \includegraphics[trim={4cm 2cm 4cm 2cm},width=0.90\columnwidth]{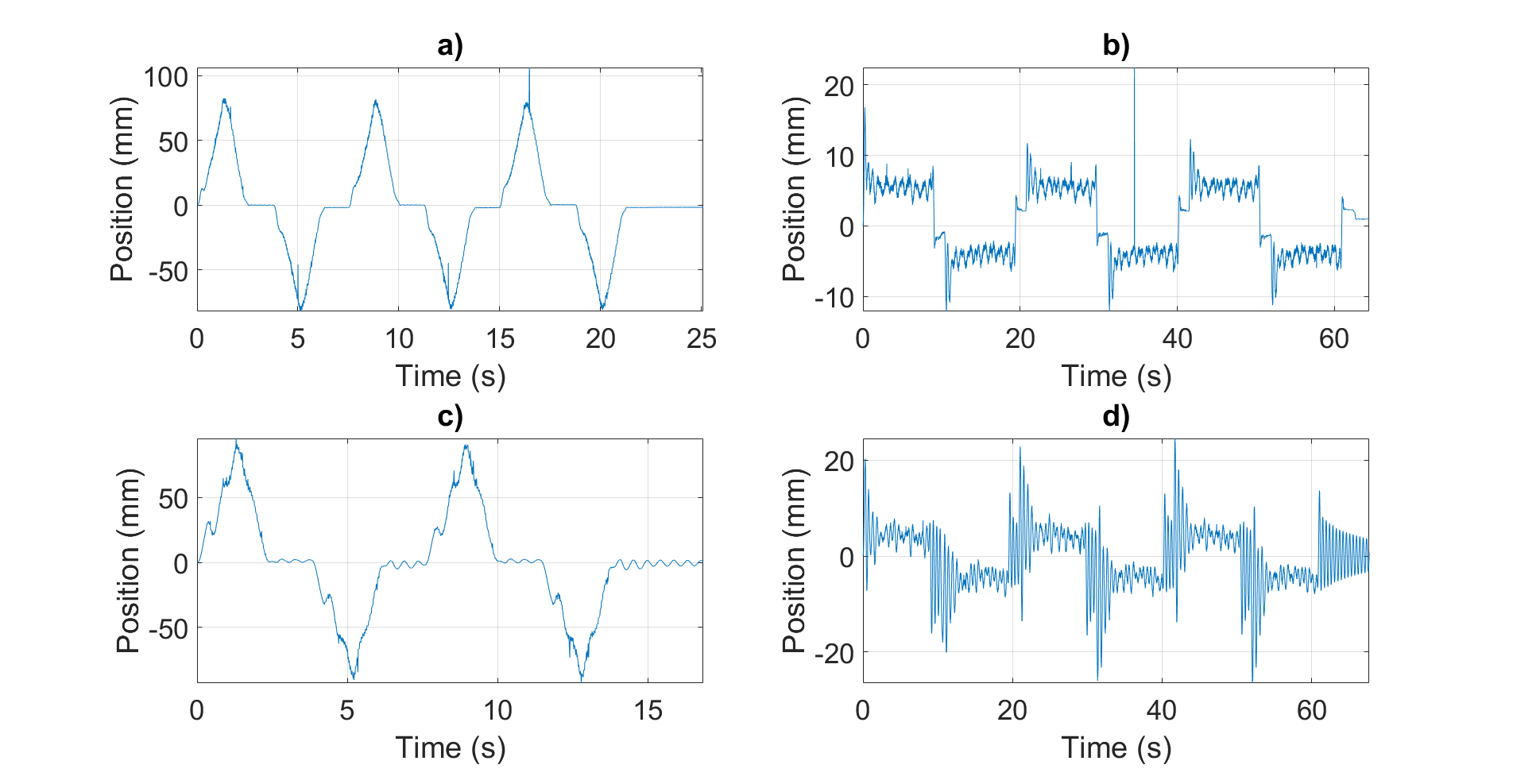}
      \caption{Error in position of the X-axis. The X-axis is driven at fast and slow speed at a high and low Z-height. (a) high-fast, (b) high-slow, (c) low-fast, (d) low-slow.}
      \label{fig:x_error}
\end{figure}

\subsection{Results}
\label{section:results}

A summary of the tests conducted can be found in Table~\ref{table:stats}. Below are the plots of that data that was captured and summarised in these tables.\\

\input{tables/stats}

\subsubsection*{X-axis}

The results of X-axis movements can be seen in Fig.~\ref{fig:x_error}. The steady can be seen to converge to close to zero once motion has ceased. The average error in position is 9.62 mm as seen in Table \ref{table:stats}. During slow movements, oscillations occur about the desired point due to the pendulum effect but still converge to the desired end point. This is considered sufficient for the application that the manipulator is designed for. \\

\begin{figure*}[t]
      \centering
      \includegraphics[width=\textwidth]{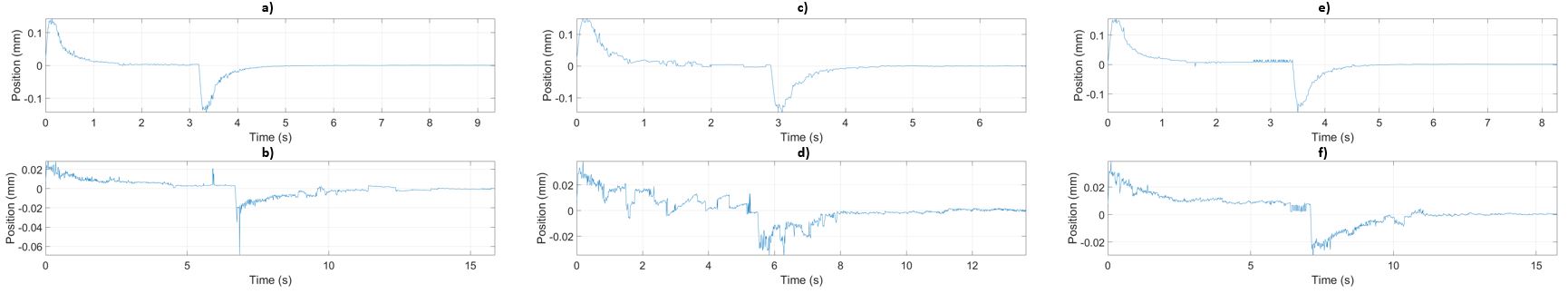}
      \caption{Wrist trajectory at fast and slow speeds. (a) roll-fast, (b) roll-slow, (c) pitch-fast, (d) pitch-slow, (e) yaw-fast, (f) yaw-slow.}
      \label{fig:wrist_error}
  \end{figure*}

 \begin{figure}[]
      \centering
      \includegraphics[trim={4cm 2cm 4cm 4cm},width=0.90\columnwidth]{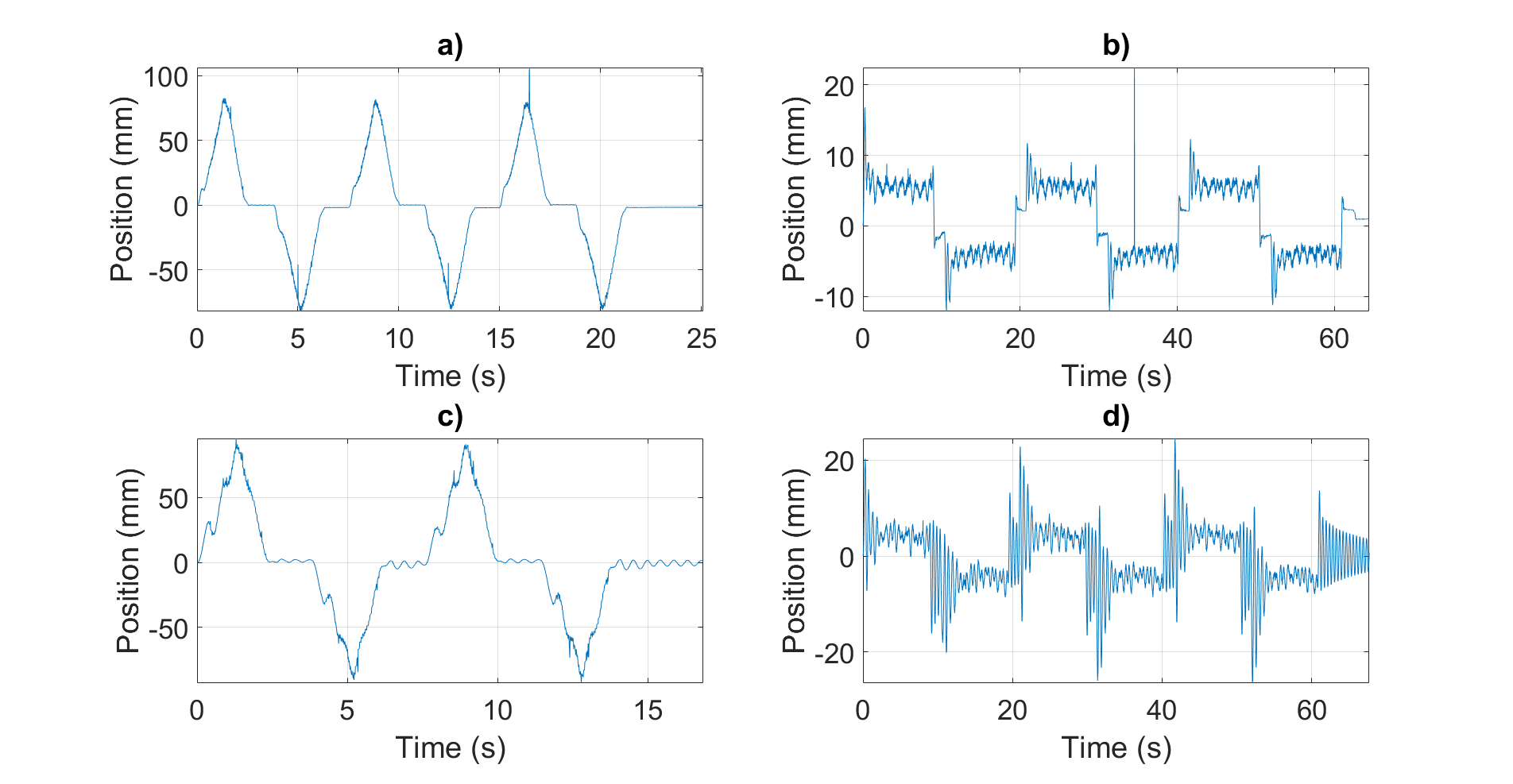}
      \caption{Error in position of the Y-axis. The Y-axis is driven at fast and slow speed at a high and low Z-height. (a) high-fast, (b) high-slow, (c) low-fast, (d) low-slow.}
      \label{fig:y_error}
   \end{figure}

 \begin{figure}[]
      \centering
      \includegraphics[trim={4cm 2cm 4cm 4cm},width=0.80\columnwidth]{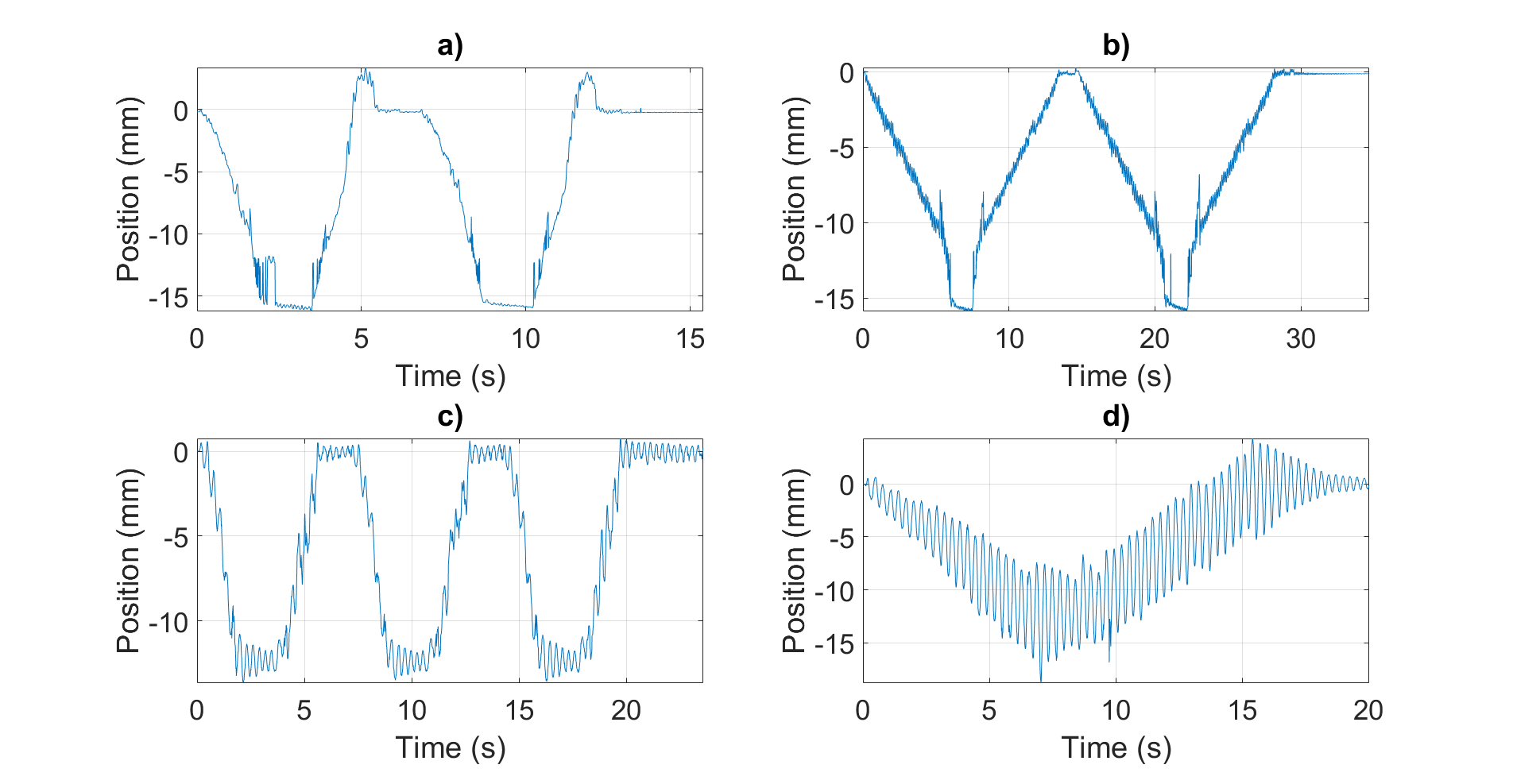}
      \caption{Z-axis trajectory during pure Y-axis motion. The oscillations are a result of the pendulum effect. (a) high-fast, (b) high-slow, (c) low-fast, (d) low-slow. \newline}
      \label{fig:yz_error}
   \end{figure}

\subsubsection*{Y-axis}

As seen in Fig.~\ref{fig:y_error}, the Y-axis trajectory oscillates substantially at low heights at slow speeds. This is caused by the pendulum effect of the system as described in Section \ref{section:design}.  For the task of the Amazon Robotics Challenge, final point accuracy is more crucial than in-motion accuracy. We can see that the final desired position converges to almost zero. Given this specific task, slow Y-axis movements are not necessary any way. 

The coupling effect of the differential belt system is also of interest. The trajectory of the Z-axis while moving the Y-axis can be seen in Fig.~\ref{fig:yz_error}. We can see that the Z-axis shows a small translation during pure Y-axis motion as a result of the differential belt system. This is to be expected as the two ClearPath motors would need to be perfectly synchronised in order to reduce this movement to zero. Despite this travel, the final point accuracy results in no unwanted Z-axis translation.\\

\subsubsection*{Z-axis}
\begin{figure}[]
      \centering
      \includegraphics[trim={4cm 2cm 4cm 4cm},width=0.90\columnwidth]{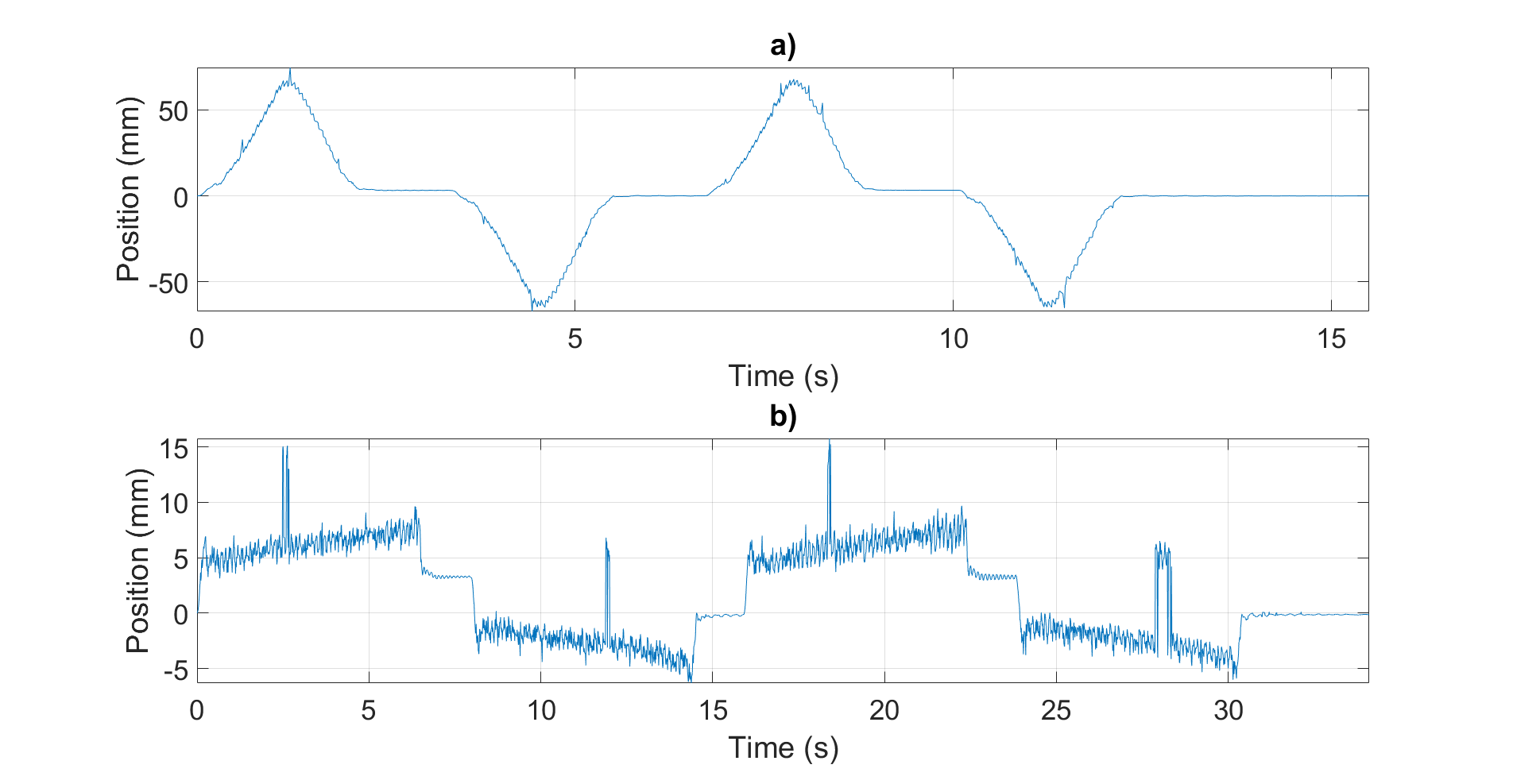}
      \caption{Z-axis trajectory at various speeds. (a) high-fast, (b) high-slow, (c) low-fast, (d) low-slow.}
      \label{fig:z_error}
   \end{figure}

The Z-axis can follow a pure Z trajectory with a fair amount of accuracy as seen in Fig.~\ref{fig:z_error}. The Z-axis exhibits the least amount of error while in motion with only 7.9 mm mean error in position. As discussed previously, due to the coupling of the Y- and Z-axes, the Y-axis wanders during pure Z-axis motion just as the Z-axis wanders during pure Y-axis motion. Despite this, steady state precision is still achieved.\\

\subsubsection*{Wrist - roll, pitch and yaw}

The wrist joint of the robot exhibit exceptional accuracy not only in motion but in steady state as well as seen in Fig.~\ref{fig:wrist_error}. Achieving a steady state of almost zero error.\\

\subsubsection*{XYZ axes}

\begin{figure}[]
      \centering
      \includegraphics[trim={4cm 2cm 4cm 4cm},width=0.90\columnwidth]{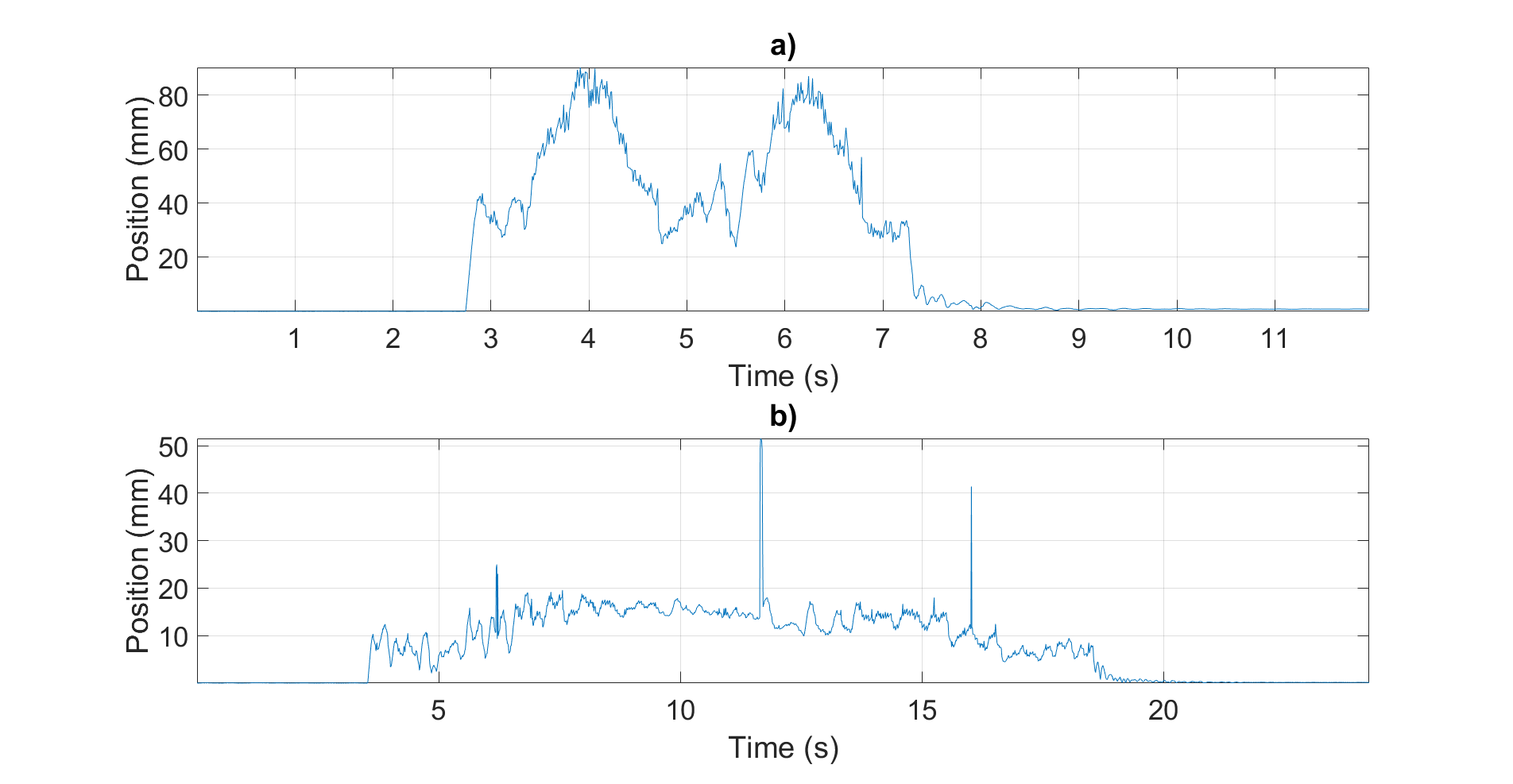}
      \caption{Three axis movements in the X-, Y-, and Z-axes. (a) xyz-fast, (b) xyz-slow.}
      \label{fig:xyz_error}
\end{figure}

Simultaneous motion of multiple axes were tested against a more complex trajectory. A lemniscate curve was generated on firstly a single plane at three different heights and also in a multi-plane test. The system was tested at different speeds and exhibited the same characteristics as the single axis. Steady state error is almost zero seen in Fig.~\ref{fig:xyz_error}.\\


\subsubsection*{Example Pick run}

\begin{figure}[]
      \centering
      \includegraphics[trim={4cm 2cm 4cm 4cm},width=0.6\columnwidth]{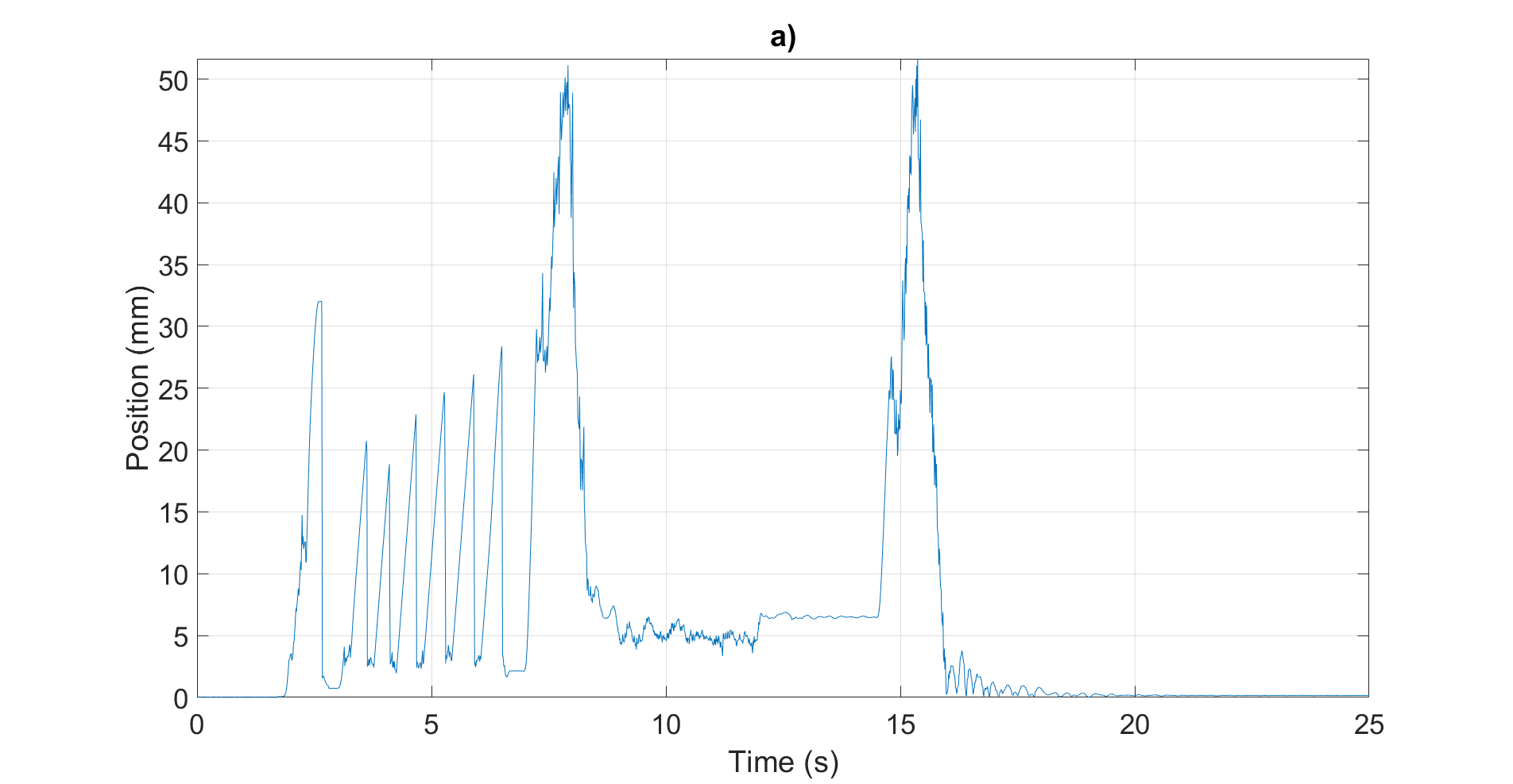}
      \caption{Example trajectory of a real pick. The task was to pick up a 1kg dumbbell from a start point, place it in another area and return to the start position}
      \label{fig:pick_error}
\end{figure}

To simulate a pick task within the Amazon Robotics Challenge a simple scenario was set up in order to test accuracy during motion the manipulator was intended for. The task was to pick an item from the Amazon item set, in this case a 2lb weight, and place it to another location and then return to its start position. The trajectory recorded can be seen in Fig.~\ref{fig:pick_error}. The manipulator was easily able to achieve the task with a mean error of only  2.96mm  during motion.\\

\subsection{Verification}
\label{section:verification}

\input{tables/speed}

Using the data presented in this section, we can verify our speed specifications. Unfortunately due to mechanical limitations, the system experienced belt slip under high accelerations. As a result the acceleration parameter of the controller needed to be lowered. Table \ref{table:speed} provides a summary of the speed and error achieved by each axis. We've shown that the system is capable of moving a load whilst following linear trajectories with reasonable accuracy. Table \ref{table:stats} features the static error for each axis. We found that despite oscillations and trajectory error during motion, the steady state error is negligible and can achieve sub-mm and sub-mrad accuracy.

%% file: tables/stats.tex
\newcolumntype{s}{>{\centering\hsize=0.9\hsize\arraybackslash}X}
\begin{table}[t]
\centering
\caption{Summary of error across single and multiple axes}
\label{table:stats}
\begin{tabularx}{0.9\columnwidth}{lsss}
\toprule
               & Standard Deviation  & Mean Error  & Static Error\\ 
\midrule
\textbf{x}     & 16.07  & 9.62  & $6.2e^{-01}$ mm  \\
\textbf{y}     & 13.73  & 9.42  & $1.9e^{-01}$ mm  \\
\textbf{z}     & 13.70  & 7.91  & $1.9e^{-03}$ mm  \\
\textbf{xzy}   & 64.88  & 29.63 & $1.5e^{-01}$ mm  \\
\textbf{roll}  & 0.02   & 0.01  & $5.6e^{-04}$ rad \\
\textbf{pitch} & 0.02   & 0.01  & $1.5e^{-04}$ rad \\
\textbf{yaw}   & 0.02   & 0.01  & $7.3e^{-05}$ rad \\
\bottomrule
\end{tabularx}
\begin{tablenotes}
      \item Each axis is driven individually and simultaneous motion of axes are also evaluated (xyz). Mean error is defined as the error between the desired position and the measured position. Static error is the error between desired position and measured position once motion has ceased. \newline
\end{tablenotes}

\end{table}


%% file: tables/speed.tex
\newcolumntype{s}{>{\centering\hsize=.5\hsize\arraybackslash}X}

\begin{table}[t]
\centering
\caption{Summary of speeds achieved by the system}
\label{table:speed}
\begin{tabularx}{0.9\columnwidth}{lccc}
\toprule
Axis       & Commanded Speed & Achieved Speed & \% Error \\
\midrule
x     & 0.604           & 0.572          & 5\%      \\
y     & 0.531           & 0.493          & 7\%      \\
z     & 0.512           & 0.472          & 8\%      \\
roll  & 1.547           & 1.687          & -9\%     \\
pitch & 1.536           & 1.531          & 0\%      \\
yaw fast   & 1.593           & 1.660          & -4\% \\ 
\bottomrule
\end{tabularx}
\begin{tablenotes}
      \item As the controller used for the system is a position controller, negative error values indicate that the joint is able to track cope with latency and accurately track a trajectory.
\end{tablenotes}
\end{table}
